\documentclass{article} %
\usepackage{colm2024_conference}

\usepackage{booktabs}
\usepackage{graphicx}
\usepackage{enumitem}
\usepackage{wrapfig}
\usepackage{algorithm}
\usepackage{algpseudocode}

\usepackage{microtype}
\usepackage{amsmath}
\usepackage{colortbl}
\usepackage[utf8]{inputenc}
\definecolor{lightgray}{rgb}{0.9,0.9,0.9}
\usepackage{caption}
\usepackage{subcaption}
\usepackage{xcolor}
\usepackage{setspace}
\usepackage{url}
\usepackage{multirow}
\usepackage{colortbl}
\usepackage{tabularx}
\usepackage{blindtext}
\usepackage{pgfplots}
\pgfplotsset{compat=1.18} 
\usepackage{tikz}
\usetikzlibrary{er,positioning,bayesnet}
\usepackage{makecell}
\usepackage{tipa}
\usepackage{siunitx}
\usepackage{nicefrac}
\usepackage{tocloft}
\usepackage{listings}
\usepackage[raster,skins]{tcolorbox} %
\usepackage{xltabular}
\usepackage{adjustbox}
\usepackage{xurl}
\usepackage{cleveref}

\usepackage{amsmath,amsfonts,bm}

\def\eqref#1{equation~\ref{#1}}

\def\1{\bm{1}}

\DeclareMathAlphabet{\mathsfit}{\encodingdefault}{\sfdefault}{m}{sl}
\SetMathAlphabet{\mathsfit}{bold}{\encodingdefault}{\sfdefault}{bx}{n}

\title{Qwen2.5-1M Technical Report}

\author{
An Yang, Bowen Yu, Chengyuan Li, Dayiheng Liu, Fei Huang, Haoyan Huang, Jiandong Jiang, Jianhong Tu, Jianwei Zhang, Jingren Zhou, Junyang Lin, Kai Dang, Kexin Yang, Le Yu, Mei Li, Minmin Sun, Qin Zhu, Rui Men, Tao He, Weijia Xu, Wenbiao Yin, Wenyuan Yu, Xiafei Qiu, Xingzhang Ren, Xinlong Yang, Yong Li, Zhiying Xu, Zipeng Zhang\thanks{Authors are ordered alphabetically by the last name.}\\
\vspace{0.5em}
\textbf{Qwen Team, Alibaba Group}
\vspace{-1.5em}
}

\begin{document}

\maketitle

\begin{abstract}

In this report, we introduce Qwen2.5-1M, a series of models that extend the context length to 1 million tokens. Compared to the previous 128K version, the Qwen2.5-1M series have significantly enhanced long-context capabilities through long-context pre-training and post-training. Key techniques such as long data synthesis, progressive pre-training, and multi-stage supervised fine-tuning are employed to effectively enhance long-context performance while reducing training costs.

To promote the use of long-context models among a broader user base, we present and open-source our inference framework. This framework includes a length extrapolation method that can expand the model context lengths by at least four times, or even more, without additional training. To reduce inference costs, we implement a sparse attention method along with chunked prefill optimization for deployment scenarios and a sparsity refinement method to improve precision. Additionally, we detail our optimizations in the inference engine, including kernel optimization, pipeline parallelism, and scheduling optimization, which significantly enhance overall inference performance. By leveraging our inference framework, the Qwen2.5-1M models achieve a remarkable 3x to 7x prefill speedup in scenarios with 1 million tokens of context. This framework provides an efficient and powerful solution for developing applications that require long-context processing using open-source models.

The Qwen2.5-1M series currently includes the open-source models Qwen2.5-7B-Instruct-1M and Qwen2.5-14B-Instruct-1M, as well as the API-accessed model Qwen2.5-Turbo. Evaluations show that Qwen2.5-1M models have been greatly improved in long-context tasks without compromising performance in short-context scenarios. Specifically, the Qwen2.5-14B-Instruct-1M model significantly outperforms GPT-4o-mini in long-context tasks and supports contexts eight times longer.

\end{abstract}
\vspace{-0.5em}

\vfill

\begin{figure}[hbp]
    \centering
    \includegraphics[width=0.9\textwidth]{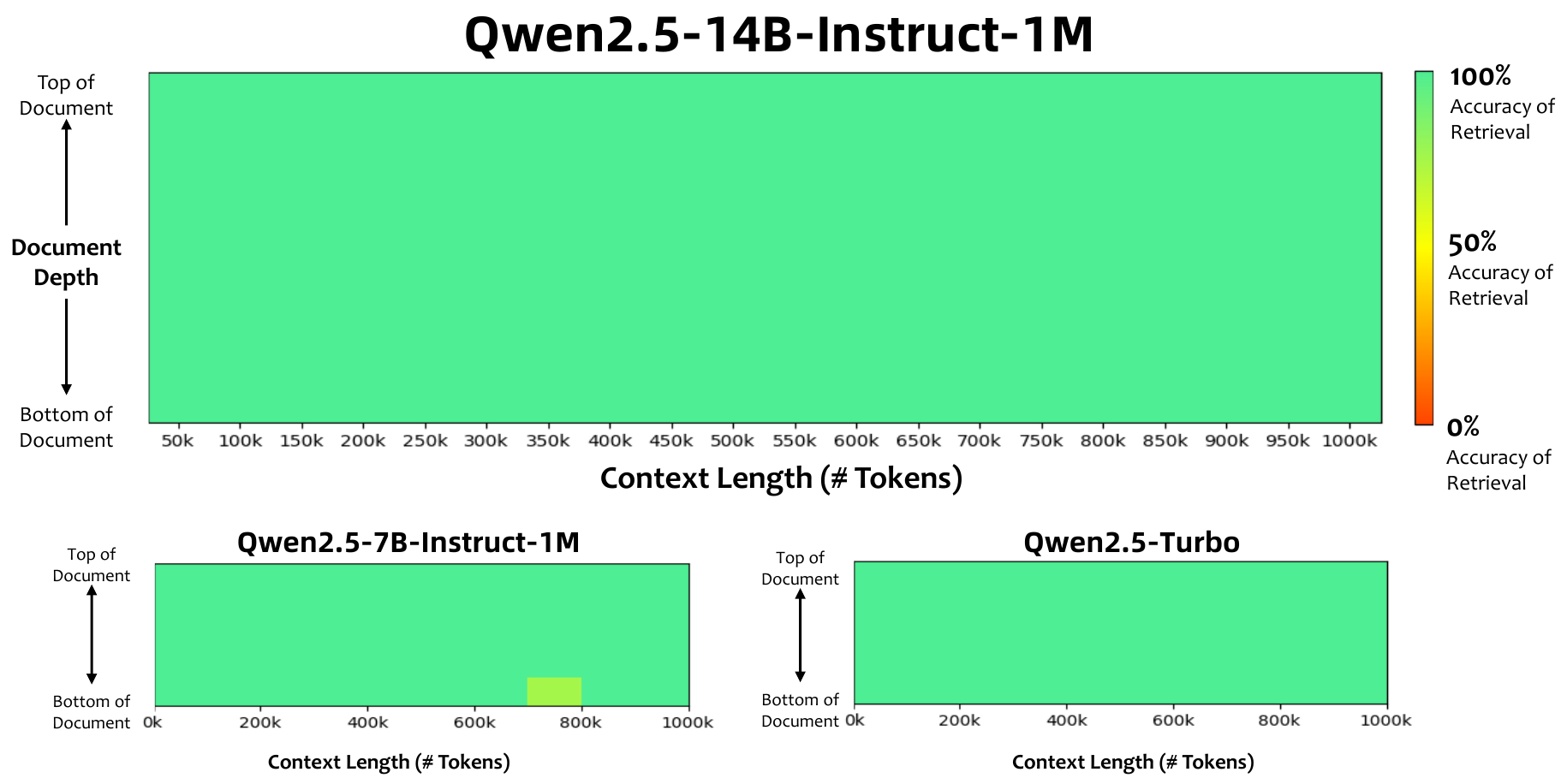}
    \caption{\textbf{Passkey Retrieval Test on Qwen2.5-1M Models with documents up to 1 Million Tokens.} This test evaluates the model's ability to retrieve a hidden number from ultra-long documents filled with irrelevant content. The results show that the Qwen2.5-1M models can accurately retrieve hidden numbers from documents containing up to 1M tokens, with only minor errors observed in the 7B model.}
     \label{fig:cover}
\end{figure}

\vfill

\newpage

\section{Introduction}
\label{sec:intro}

Large Language Models (LLMs) have revolutionized natural language processing by demonstrating remarkable capabilities in understanding, generating, and interacting with human language~\citep{gpt3,gpt4,gpt4o,gemini,claude,claude2,claude3,qwen,qwen2,qwen2.5,llama,llama2,llama3, mistral, mixtral}. However, the limited context length restricts the amount of text that they can process at once, confining their capabilities to simpler, single tasks and preventing them from tackling complex real-world scenarios that require extensive information processing or generation. For example, LLMs struggle with performing code generation and debugging that rely on repository-level context or conducting in-depth research based on large volumes of documents. 

To address this, increasing the context window of LLMs has become a significant trend. Models like GPT series~\citep{gpt3, gpt4, gpt4o}, LLama series~\citep{llama, llama2, llama3}, and our Qwen series~\citep{qwen, qwen2, codeqwen, qwen2.5coder, qwen2math, qwen2.5math} have rapidly expanded from initial context windows of 4k or 8k tokens to the current 128k tokens. There are also explorations to extend the context length of LLMs to 1M tokens or even longer, such as Gemini~\citep{gemini}, GLM-9B-Chat-1M~\citep{chatglm4}, and Llama-3-1M models from Gradient AI~\citep{gradientlongcontextllama3}. This growth has enabled more sophisticated applications, allowing both users and developers to leverage these models' enhanced context capabilities for innovative research and development.

In this report, we will introduce the 1M context length version of Qwen2.5, namely the Qwen2.5-1M series. In terms of open-source weights, we release two instruction-tuned models: Qwen2.5-7B-Instruct-1M and Qwen2.5-14B-Instruct-1M. Compared to the 128K versions, these models exhibit significantly enhanced long-context capabilities. Additionally, we provide an API-accessible model based on Mixture of Experts (MoE), called Qwen2.5-Turbo, which offers performance comparable to GPT-4o-mini but with longer context, stronger capabilities, and more competitive pricing.
Beyond the models themselves, we also open-source our inference framework optimized for long-context processing, enabling developers to deploy the Qwen2.5-1M models more cost-effectively.

This report outlines the key methodologies behind Qwen2.5-1M, focusing on two main aspects:

\begin{itemize}
    \item \textbf{Efficient Long-Context Training.} 
    The pre-training of Qwen2.5-1M incorporates synthetic data emphasizing long-range dependencies, with a progressive length extension strategy to reduce costs and enhance efficiency. Post-training addresses the scarcity of long-instruction datasets using agent-generated large-scale instruction data. A multi-stage Supervised Fine-Tuning (SFT) and Reinforcement Learning (RL) ensures balanced performance across short and long sequences, optimizing alignment with human preferences.

    \item \textbf{Efficient Inference and Deployment.} Our inference framework encompasses three key components: (1) a training-free length extrapolation method that allows models trained on 256k context lengths to seamlessly scale up to 1M contexts without requiring additional training; (2) a sparse attention mechanism aimed at reducing inference costs, with further optimizations to enhance GPU memory efficiency, integration with the length extrapolation method, and refined sparsity configurations to boost accuracy; and (3) engine-level optimizations such as kernel improvements, pipeline parallelism, and enhanced scheduling. By leveraging these advancements, our inference framework boost prefill speeds by 3 to 7 times in 1M-context scenarios.
    
\end{itemize}

\section{Architecture}

Qwen2.5-1M series are developed based on Qwen2.5 models~\citep{qwen2.5} and support context length up to 1M tokens. It currently includes two dense models for opensource, namely Qwen2.5-7B-1M, Qwen2.5-14B-1M, and a MOE model for API service, namely Qwen2.5-Turbo.

The Qwen2.5-1M models retain the same Transformer-based architecture as Qwen2.5, ensuring compatibility in inference. Specifically, the architecture incorporates Grouped Query Attention (GQA, \citealp{gqa}) for efficient KV cache utilization, the SwiGLU activation function~\citep{glu} for non-linear transformations, Rotary Positional Embeddings (RoPE, \citealp{rope}) to encode positional information, QKV bias~\citep{qkv_bias} in the attention mechanism, and RMSNorm~\citep{rmsnorm} with pre-normalization to ensure stable training.

\begin{table}[h]
\caption{\textbf{Model architecture and license of Qwen2.5-1M open-weight models.}}
\small
\centering
\begin{tabular}{@{}lcccccccc@{}} 
\toprule
Models  & Layers & Heads (Q / KV) & Tie Embedding & Context / Generation Length  & License \\
\midrule
7B  & 28 & 28 / 4 & No & 1M / 8K & Apache 2.0 \\
14B & 48 & 40 / 8 & No & 1M / 8K & Apache 2.0 \\
\bottomrule
\end{tabular}
\end{table}

\section{Pre-training}

Long-context pre-training is computationally intensive and can be expensive. To enhance training efficiency and reduce costs, we focus on optimizing data efficiency and refining training strategies during the pre-training process of Qwen2.5-1M models. Specifically, our improvements come from the following aspects:

\paragraph{Natural and Synthetic Data.}

During the pre-training phase, we assemble an extensive and diverse corpus of natural long-text data to ensure that our Qwen2.5-1M models are exposed to a wide array of linguistic patterns and contexts. This corpus encompasses various domains, including but not limited to Common Crawl, arXiv, books, and code repositories.

Despite the richness, natural corpus often exhibits weak long-distance associations, making it challenging for models to learn the connections between distant tokens effectively. This limitation arises because natural texts typically prioritize local coherence over global structure, where the model can effortlessly predict the next token without relying on long-range dependencies.

To address these challenges, we augmented the natural corpus with synthetic data designed to enhance the model's capacity to understand and generate long-range dependencies. The synthetic data generation process involved several sophisticated tasks aimed at improving the model's comprehension of sequential relationships and contextual understanding:

\begin{itemize}
    \item \textbf{Fill in the Middle} (FIM, \citealp{fim2022openai}): FIM tasks require the model to predict missing segments within a given text sequence. By inserting gaps at various positions and lengths, FIM encourages the model to focus on integrating distant contextual information surrounding the gap.
    \item \textbf{Keyword-Based and Position-Based Retrieval}: This task involves retrieving relevant paragraphs based on specific keywords or recalling paragraphs that appear before or after a specified position. This task enables the model to enhance its ability to identify and connect relevant information across different parts of a text while improving its understanding of positional relationships within sequences.    
    \item \textbf{Paragraph Reordering}: In this task, paragraphs are shuffled, and the model must reorder them to restore the original sequence. This task strengthens the model's ability to recognize logical flows and structural coherence, essential for generating well-organized and coherent text.
\end{itemize}

By integrating these synthetic data tasks into the pre-training process, we significantly improved the model's ability to capture long-range information. This approach not only enhances data efficiency but also reduces the overall computational cost by accelerating the learning process and requiring fewer iterations to achieve high performance.

\paragraph{Training Strategy.} Training with long contexts requires substantial GPU memory, thus posing a severe challenge to both training costs and time. To improve training efficiency, the Qwen2.5-1M models adopted a progressive context length expansion strategy, which includes five stages.

The first two stages are similar to those of other Qwen2.5 models, where we directly use an intermediate version from Qwen2.5 Base models for subsequent long-context training. Specifically, the model is initially trained with a context length of 4096 tokens, and then the training is transferred to a context length of 32768 tokens. During this process, we employ the Adaptive Base Frequency (ABF) technique~\citep{ropeabf}, adjusting the base frequency of the Rotary Position Embedding (RoPE, \citealp{rope}) from 10,000 to 1,000,000.

In the subsequent three stages, the context lengths are expanded to 65,536 tokens, 131,072 tokens, and 262,144 tokens, with the RoPE base frequencies set to 1,000,000, 5,000,000, and 10,000,000, respectively. During these stages, the training data is curated to include 75\% sequences at the current maximum length and 25\% shorter sequences. This approach ensures that the model can effectively adapt to longer contexts while preserving its capability to process and generalize across sequences of different lengths.

\begin{table}[tbp]
\centering
\caption{\textbf{Performance of Qwen2.5-14B-1M on RULER at each pre-training stage.}}
\label{tab:pretrain_stage}
\small
\begin{tabular}{@{}rlllllll@{}}
\toprule
\multirow{2}[2]{*}{\makecell{\bf Training Length}} & \multicolumn{6}{c}{\bf RULER}  \\ 
\cmidrule{2-8}
 & \bf Avg.  & \bf 4K   & \bf 8K    & \bf 16K  & \bf 32K  & \bf 64K   & \bf 128K  \\ \midrule
32,768 Tokens   & 82.3 & 96.8 & 94.7 & 95.9 & 92.2 & 76.4 & 37.6   \\
65,536 Tokens   & 86.8 & 96.5 & 95.5 & 93.6 & 92.5 & 86.7 & 56.0 \\
131,072 Tokens & 92.5 & 96.5 & 95.9 & 93.0 & 92.6 & 93.0 & 83.8 \\
262,144 Tokens  & 92.7 & 95.6 & 93.8 & 93.1 & 94.1 & 91.9 & 87.6  \\ 
\bottomrule
\end{tabular}
\end{table}

To monitor the performance changes of the progress training, we evaluate Qwen2.5-14B-1M on the RULER~\citep{hsieh2024ruler} benchmark at the end of each training phase. As illustrated in Table \ref{tab:pretrain_stage}, training with progressively longer sequences consistently enhances the model's comprehension capabilities for the corresponding sequence lengths.
Notably, even the final pre-training stage, which uses sequences of 262,144 tokens, significantly improves performance on the 128K samples. This finding aligns with previous research \citep{an2024string}, suggesting that models benefit significantly from training on longer sequences to fully realize their potential on relatively shorter tasks.

\section{Post-Training}

The aim of post-training is to effectively enhance the model's performance on long-context tasks while ensuring that performance on short tasks does not decline. We highlight the following efforts in building the Qwen2.5-1M models during post-training:

\paragraph{Synthesizing Long Instruction Data.} In long-context tasks, human annotation can be expensive and unreliable. To address this issue, our training data includes a substantial portion of synthetic long-context question-answer pairs. Specifically, inspired by \cite{llama3, longalign2024bai}, we select long documents from the pre-training corpus and prompt Qwen2.5 to generate queries based on a randomly extracted segment of each document. These queries encompass a variety of tasks, including summarization, information retrieval, multi-hop question answering, reasoning, coding, and others. We then leverage the Qwen-Agent framework~\citep{qwen-agent-2405} to generate high-quality responses based on the full documents. This framework employs advanced techniques such as retrieval-augmented generation, chunk-by-chunk reading, and step-by-step reasoning, enabling it to integrate the overall content of the documents into its responses comprehensively. Finally, we utilize the full documents, the model-generated queries, and the agent-based generated responses to constitute synthetic training data.

\paragraph{Two-stage Supervised Fine-tuning.} To enhance the model's performance on long-context tasks without compromising its performance on shorter tasks, we utilize a two-stage training scheme.
In the first stage, similar to the Qwen2.5 models, we trained the model exclusively on short instruction data, each containing up to 32,768 tokens, and maintained the same number of training steps. This stage ensures that the model retained its proficiency in handling short tasks.
In the second stage, we introduce a mixed dataset comprising both short and long sequences, with lengths ranging from up to 32,768 tokens to up to 262,144 tokens. We carefully balance the ratio of short to long data to prevent the model from forgetting the skills it has acquired during the first stage.

\paragraph{Reinforcement Learning.} We employ offline reinforcement learning, similar to Direct Preference Optimization (DPO, \citealp{rafailov2024direct}), to enhance the model's alignment with human preferences. Specifically, we utilize the training pairs from the offline RL phase of other Qwen2.5 models, which consisted solely of short samples up to 8,192 tokens. We find that training on these short samples is sufficient to significantly improve the model's alignment with human preferences and to generalize effectively to long-context tasks. 
To substantiate this claim, we evaluated the model before and after RL using the longbench-chat benchmark. As shown in Table \ref{tab:rl_ablation}, the RL stage lead to significant improvements across all models, demonstrating the effective generalization of RL from short-context to long-context tasks.

\begin{table}[h]
\centering
\caption{\textbf{Performance on Longbench-Chat before and after RL stage.}}
\label{tab:rl_ablation}
\small
\begin{tabular}{lcc}
\toprule
Model & Before RL & After RL \\
\midrule
Qwen2.5-7B-Instruct-1M & 7.32 & 8.08 (+0.75) \\
Qwen2.5-14B-Instruct-1M & 8.56 & 8.76 (+0.20) \\
Qwen2.5-Turbo & 7.60 & 8.34 (+0.74) \\
\bottomrule
\end{tabular}
\end{table}

\section{Inference and Deployment}

Inference and deployment present significant challenges when LLMs process long-context tasks. Key issues include deploying models with longer sequences within the constraints of limited GPU memory, reducing computation to speed up processing, while maintaining accuracy during the optimization. In this section, we will introduce our approaches to addressing these challenges.

First, we present our length extrapolation method, which enables the model to support context lengths that are four times or even greater than the training length during inference. Next, we introduce a sparse attention mechanism that achieves more than a four-fold acceleration in the prefill stage. Finally, we delve into our optimizations at both the kernel and system levels, which further enhance overall inference performance.

Our inference and deployment solution detailed in this section has been open-sourced and integrated into vLLM~\citep{vllm}. It empowers users to deploy the Qwen-2.5 model on their own devices, leveraging our advanced length extrapolation methods and acceleration optimizations for enhanced performance.

\subsection{Length Extrapolation}

Length extrapolation is an inference technique designed to enhance model performance when processing long inputs that exceed the context length used during training. We employ the following two methods to achieve length extrapolation.

\begin{figure}[t]
    \centering
    \includegraphics[width=0.97\textwidth]{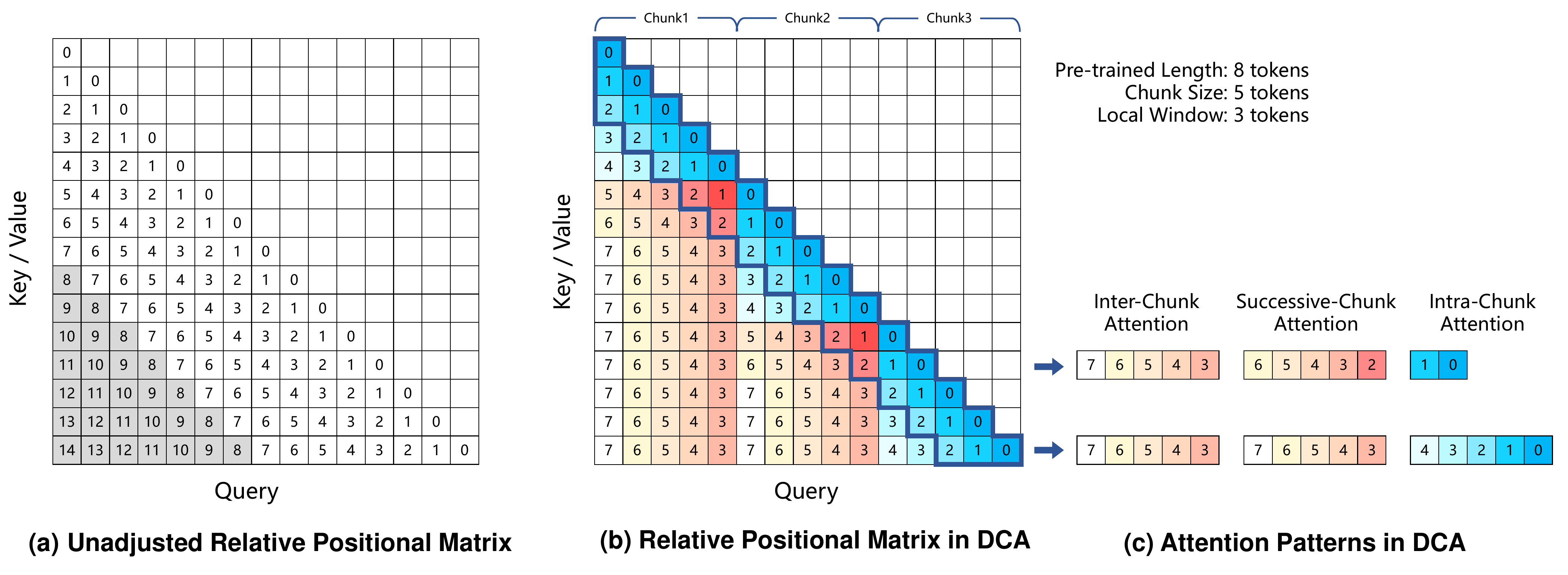}
    \caption{\textbf{An illustration of Dual Chunk Attention (DCA).} DCA remaps the relative positions to smaller numbers, thereby avoiding large relative positions that were not encountered during training (the gray areas of Figure (a)).}
     \label{fig:dca}
\end{figure}

\paragraph{Dual Chunk Attention (DCA, \citealp{chunkllama}).}

Modern LLMs based on RoPE experience performance degradation when processing sequences longer than the length in training, mainly due to encountering untrained, large relative positional distances between queries and keys in computing attention weights.

The DCA method addresses this issue by dividing the entire sequence into multiple chunks and remapping the relative positions into smaller numbers, ensuring that the distance between any two tokens does not exceed the pre-training length. An example of the remapped relative positional matrix is shown in Figure~\ref{fig:dca}(b).

DCA employs three distinct attention patterns to efficiently manage token interactions at various distances:

\begin{itemize}
    \item \textbf{Intra-Chunk Attention} handles the attention between tokens within the same chunk. Given that the distance between two tokens is relatively short, it preserves the original relative positions.
    \item \textbf{Inter-Chunk Attention} manages the attention between tokens that are not in the same chunk. To ensure that the maximum distance does not exceed the pre-training length, it uses a repeated sequences as relative positions among different chunks.
    \item \textbf{Successive-Chunk Attention} ensures the continuity of short-range relative positions by carefully managing the attention between two adjacent chunks. If the distance between a query and a key is within the local window size, it retains the original relative positions. Otherwise, it adopts the approach used in Inter-Chunk Attention to handle longer distances.
\end{itemize}

By integrating these patterns, DCA enhances the model's capability to process context lengths that are four times longer or even more. Moreover, DCA can be seamlessly integrated with flash attention, and thus efficiently implemented in a production environment.

\paragraph{Attention Scaling in YaRN~\citep{yarn}.}

When processing very long sequences, the attention mechanisms in LLMs can be distracted, leading to less focused on the key information. \citet{yarn} demonstrate that introducing a temperature parameter $t$ to the attention logits can significantly enhance model performance in a simple yet effective manner. Specifically, the computation of attention weights are modified into
\begin{equation}
\text{softmax}\left( \frac{\bf{q}^T\bf{k}}{t \sqrt{D}} \right), \ \ \text{where}\ \ \sqrt{\frac{1}{t}} = 0.1 \ln (s) + 1,
\end{equation}
$\bf{q}$ and $\bf{k}$ represent the query and key vectors, respectively. The scaling factor $s$ is the ratio of the inference length to the training length, and $D$ denotes the dimension of each attention head.

In the experiments in this report, we always use attention scaling in YaRN together with DCA. Note these two length extrapolation methods do not alter the model's behavior when processing short sequences, thus ensuring that performance on shorter tasks remains unaffected.

\paragraph{Effects of Length Extrapolation}

To demonstrate the effectiveness of the length extrapolation method, we evaluate the performance of the Qwen2.5-1M models and their 128k counterparts, both with and without DCA, under a context length of 1 million tokens. We select three tasks from RULER \citep{hsieh2024ruler} for this evaluation: Passkey Retrieval, NIAH (Needle in a Haystack) with multiple queries, and NIAH with multiple values.

\begin{figure}[!h]
    \centering
    \includegraphics[width=1\textwidth]{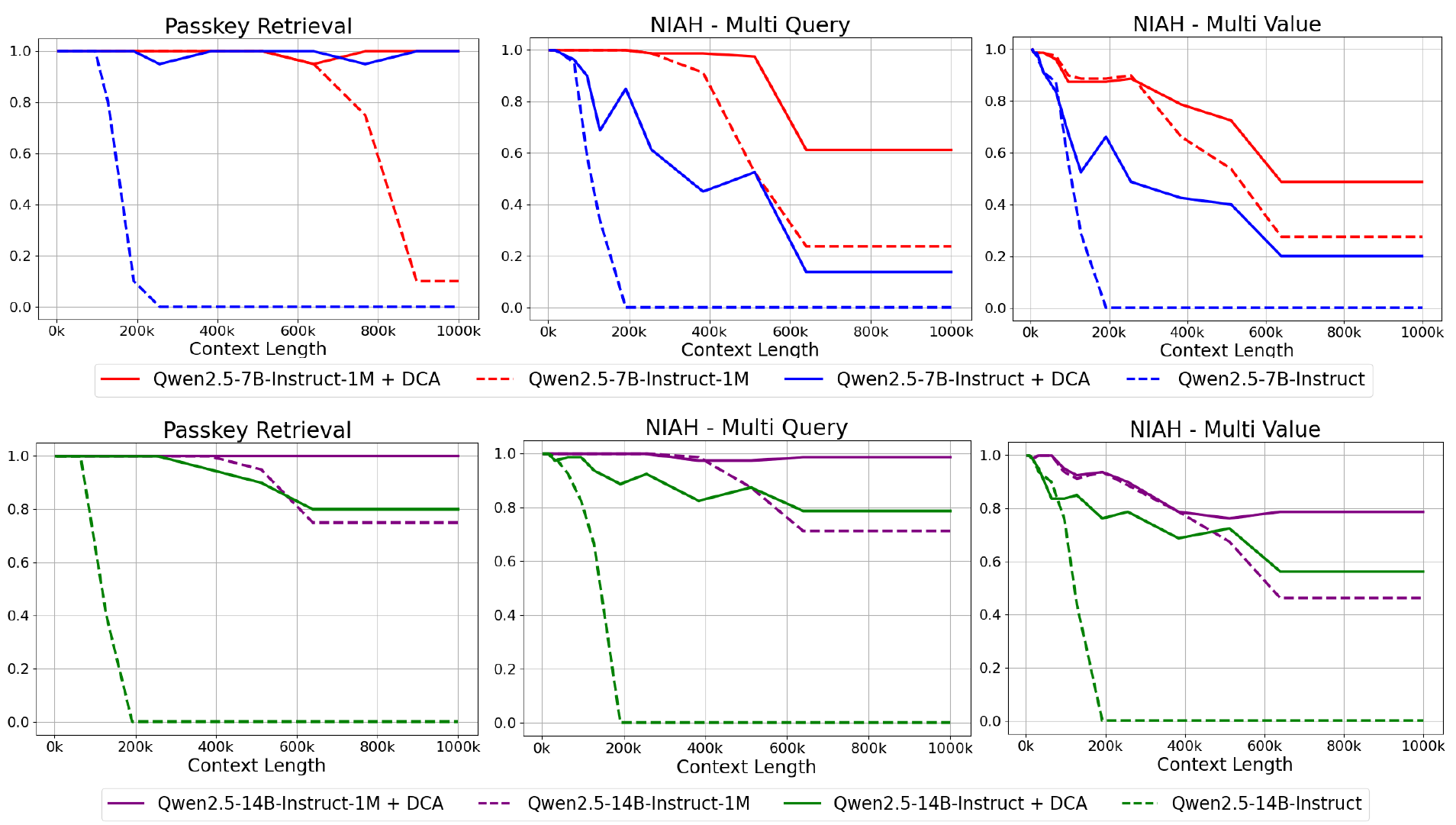}
    \caption{\textbf{The effects of length extrapolation on Long-Context Tasks.}}
     \label{fig:dca_ablation}
\end{figure}

The results are shown in Figure \ref{fig:dca_ablation}. First, we find that DCA significantly enhances the performance of all instruction models when handling long-context tasks, particularly when the context length far exceeds the length in training.
Second, for the relatively simple Passkey Retrieval task, DCA enable both the Qwen2.5-7B-Instruct and Qwen2.5-14B-Instruct models to achieve over 80\% accuracy on sequences up to 1 million tokens, despite being trained only on sequences up to 32K tokens. This underscores the efficacy of DCA as a robust solution for length extrapolation.
Finally, comparing the Qwen2.5-1M models with their 128k versions, we observed that training on longer sequences (up to 256k tokens) substantially improves the model's ability to extrapolate performance to even longer contexts.

\subsection{Efficient Inference with Sparse Attention}

For long-context LLMs, inference speed is critical to user experience. The computational complexity of conventional attention mechanisms scales quadratically with the length of the input sequence. When the input length reaches one million tokens, the time spent on the attention mechanism can account for over 90\% of the total forward pass time. Therefore, introducing sparse attention mechanisms is an essential step for the successful deployment of long-context models.

Specifically, we implemente a sparse attention mechanism based on MInference~\citep{jiang2024minference} to accelerate the prefill phase. Building on this foundation, we further optimize memory usage by integrating chunked prefill, combine these improvements with length extrapolation techniques, and introduce a sparse refinement method to address potential accuracy degradation in long sequences.

\begin{figure}[tb]
    \centering
     \begin{subfigure}[b]{0.45\textwidth}
         \centering
         \includegraphics[width=0.72\textwidth]{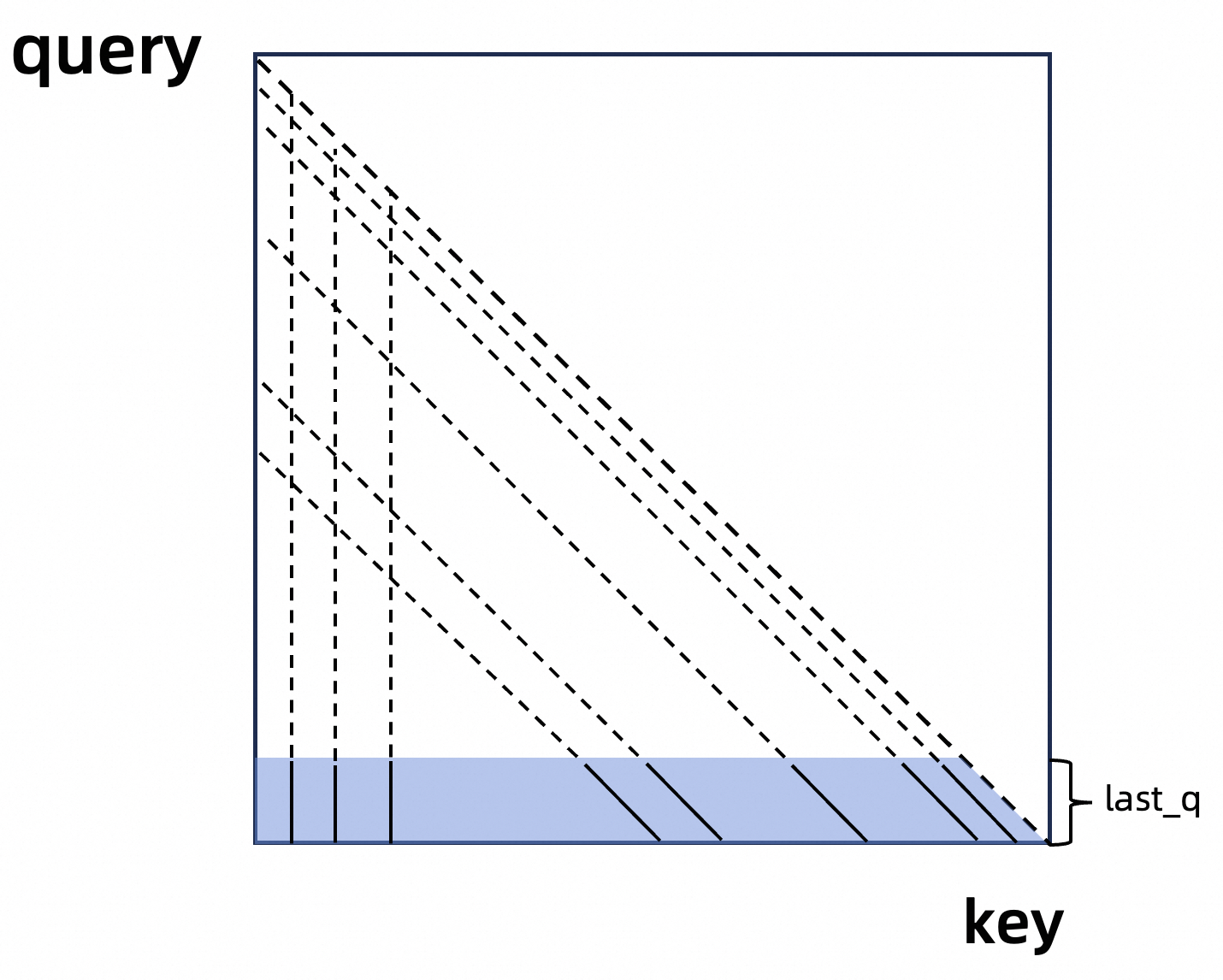}
         \caption{Vertical-Slash pattern in MInference.}
         \label{subfig:vertical_slash}
     \end{subfigure}
    ~
     \begin{subfigure}[b]{0.45\textwidth}
         \centering
         \includegraphics[width=0.72\textwidth]{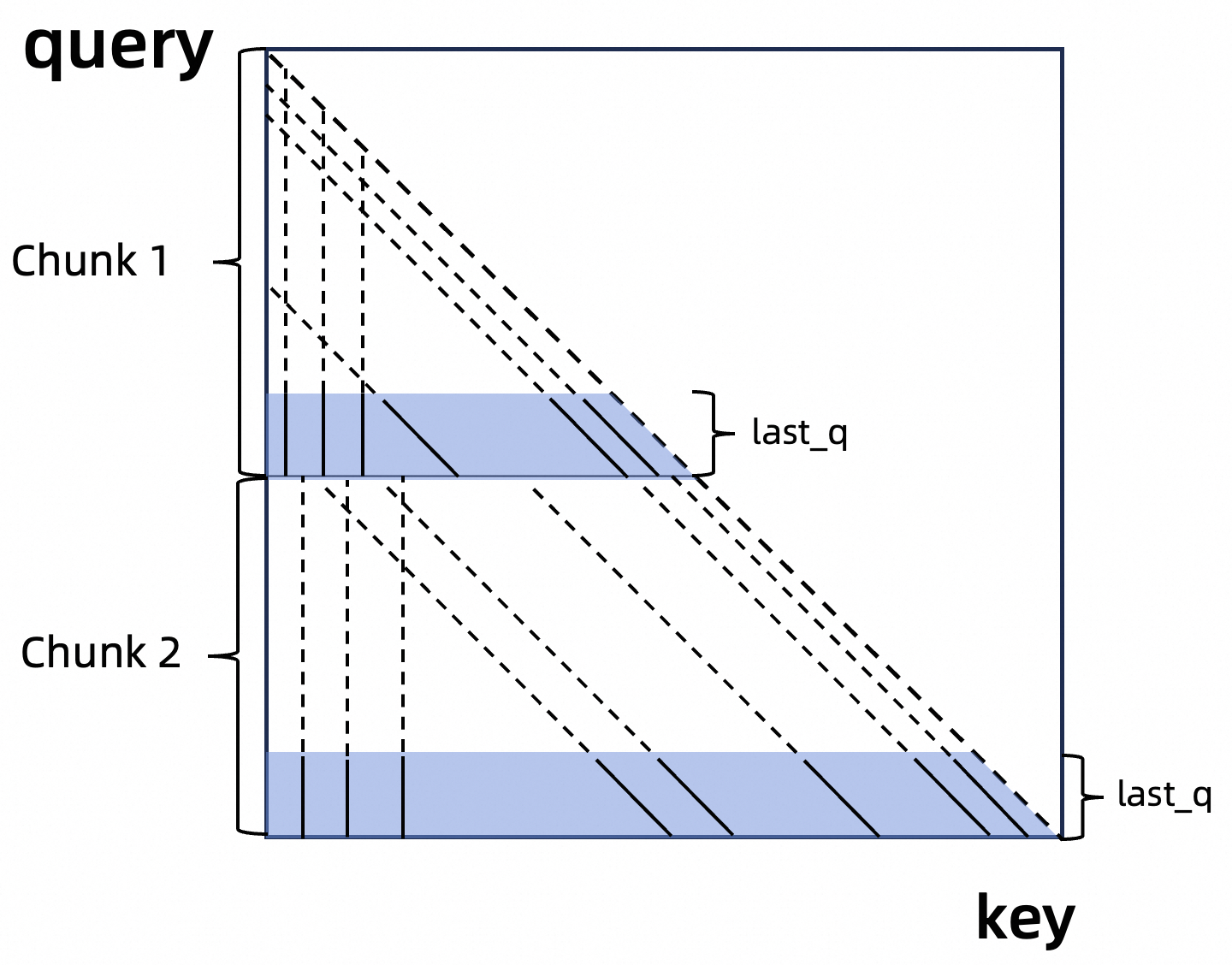}
         \caption{Combining MInference with chunked prefill.}
         \label{subfig:chunk_prefill}
     \end{subfigure}
    \caption{\textbf{An illustration of MInference and our version integrated with chunked prefill.}\label{fig:minference}}
\end{figure}

\paragraph{MInference~\citep{jiang2024minference}.}

Attention computations in large language models (LLMs) exhibit sparsity for long context inputs. \citet{jiang2024minference} successfully identify and utilize only the critical tokens for attention computation, achieving results that are nearly identical to those obtained using full attention mechanisms.
These critical tokens exhibit a distinct pattern across all samples, appearing as vertical and diagonal lines in the attention map. This pattern, referred to as the "Vertical-Slash" pattern, is illustrated in Figure \ref{fig:minference}(a).

To leverage this sparsity, MInference first conducts an offline search to determine an optimal sparsification configuration. This configuration specifies how many vertical and diagonal lines each attention head should adopt. During inference, MInference initially computes the attention between the last query tokens (i.e., last\_q) and all key tokens. Based on the partial attention results, it dynamically selects critical tokens following the "Vertical-Slash" pattern based on the pre-determined configuration, and finally computes attention only on these selected critical tokens.
This approach significantly reduces computational and memory access costs by approximately 10 times while introducing only minimal accuracy loss.

\paragraph{Integrating with Chunked prefill.}

In MInference, the entire sequence is encoded simultaneously, leading to VRAM consumption by activation values that scales linearly with the input length. For instance, when the input reaches 1 million tokens, the VRAM consumption of activation values in a single MLP layer of Qwen2.5-7B can soar to 71GB, significantly exceeding the memory usage of model weights and key-value caches.

To address this challenge, chunked prefill can be employed during inference to reduce VRAM consumption. By using a chunk length of 32,768 tokens, chunked prefill can decrease activation VRAM usage by 96.7\%. Additionally, when handling multiple requests, chunked prefill helps prevent decoding from being bottlenecked by lengthy prefill operations.

To integrate chunked prefill into MInference, we propose a strategy that selects critical tokens for each chunk, as illustrated in Figure \ref{fig:minference}(b). The input sequence is divided into multiple chunks, which are processed sequentially by the model. In the attention layer, rather than considering the last tokens of the entire input sequence that are not yet accessible, we leverage the last 64 tokens within each chunk to identify the critical tokens. This approach introduces distinct vertical and diagonal lines for each chunk during the token selection process, without significant loss in accuracy during our pilot experiments.

By integrating chunked prefill with MInference, our method significantly increases the maximum supported sequence length within limited VRAM resources.

\paragraph{Integrating with DCA.}

MInference can seamlessly integrate DCA into its implementation. However, we observe a performance drop in certain cases involving length extrapolation.

\begin{figure}[t]
    \centering
    \includegraphics[width=0.75\textwidth]{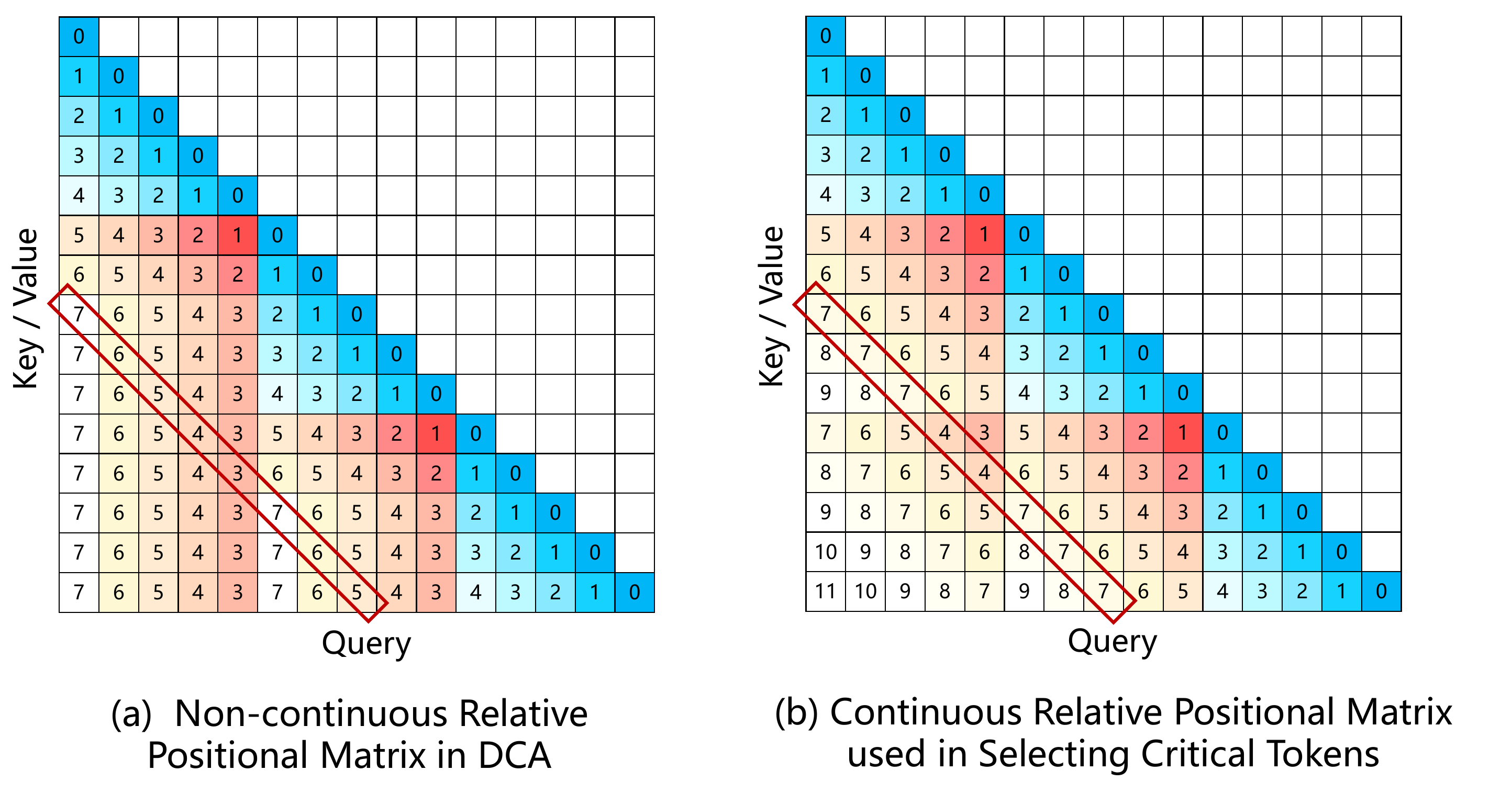}
    \caption{\textbf{Comparison of Relative Positional Matrices in DCA and in Selecting Critical Tokens.} The relative positions along the diagonal lines in (b) are more consistent than those in (a). For example, the diagonal line marked in red boxes contains $[5,6,7,3,4,5,6,7]$ in (a) and $[7,7,7,4,4,7,7,7]$ in (b).}
     \label{fig:minference_dca}
\end{figure}

We hypothesize that the non-continuity of relative positions in DCA may disrupt the ``slash'' pattern, leading to decreased accuracy in selecting critical tokens. To address this issue, we propose to recover continuous relative positions when selecting critical tokens for both successive and inter-chunk attentions, ensuring that the relative positions along the diagonal lines are as consistent as possible, as illustrated in Figure \ref{fig:minference_dca}.
It is important to note that continuous relative positions are only introduced during the critical token selection phase, and the final computation of attention weights still uses the non-continuous position embeddings in DCA.

\paragraph{Sparsity refinement on 1M sequences.} 

Before deployment, MInference requires an offline search to determine the optimal sparsification configuration for each attention head. This search process is conducted on short sequences due to the computational demand of full attention matrices, which scale quadratically with sequence length. Given the VRAM limitations of the devices used, the sequences in this search are typically kept below 32k tokens, leading to suboptimal performance on longer sequences, such as those with 1M tokens.

To address this limitation, we developed a method to refine the sparsification configuration specifically for sequences up to 1M tokens. Our approach leverages the efficient implementation of Flash Attention~\citep{flashattn} to obtain \textit{softmax\_lse}, which is defined as:
\begin{equation}
\text{softmax\_lse}_{\text{full}} = \log \sum_{0 \leq j \leq i} \exp\left(\frac{\mathbf{q}^T k_j}{\sqrt{D}}\right).
\end{equation}
The above \textit{softmax\_lse} represents the sum of unnormalized attention weights for the query $\mathbf{q}$. Similarly, we define the \textit{softmax\_lse} for sparse attention as:
\begin{equation}
\text{softmax\_lse}_{\text{sparse}} = \log \sum_{j \in \text{Critical}} \exp\left(\frac{\mathbf{q}^T k_j}{\sqrt{D}}\right),
\end{equation}
indicating the sum of unnormalized attention weights for the query $q_i$ only on critical tokens. Consequently, we can calculate the recall of attention weights as:
\begin{equation}
\text{Attention\_Recall} := \exp(\text{softmax\_lse}_{\text{sparse}} - \text{softmax\_lse}_{\text{full}}),
\end{equation}
which is between 0 and 1, indicating how well the critical tokens are captured in the sparse attention computation.

Using this attention recall metric, we refine the sparsification configuration on a calibration set consisting of 1M-token sequences. The refinement process is detailed in Algorithm \ref{alg:refinement}.

\begin{algorithm}[H]
\caption{Sparsity Refinement\label{alg:refinement}}
\begin{algorithmic}[1] 
    \Procedure{}{} 
        \For{$l \gets 1$ \textbf{to} $num\_layers$}
            \For{$h \gets 1$ \textbf{to} $num\_heads$}
                \State $\mathbf{q} \gets \text{Queries of Layer } l \text{ and Head } h$
                \State $\mathbf{k} \gets \text{Keys of Layer } l \text{ and Head } h$
                \State $\mathbf{v} \gets \text{Values of Layer } l \text{ and Head } h$
                \State $\mathbf{o}_{\text{full}},\ \text{softmax\_lse}_{\text{full}} \gets \text{full\_attention}(\mathbf{q}, \mathbf{k}, \mathbf{v})$
                \\
                \State $c \gets \text{Sparsity configs on Layer } l \text{ and Head } h$
                \State $\mathbf{o}_{\text{sparse}},\ \text{softmax\_lse}_{\text{sparse}} \gets \text{sparse\_attention}(\mathbf{q}, \mathbf{k}, \mathbf{v}, c)$
                \State \If{$\exp(\text{softmax\_lse}_{\text{sparse}} - \text{softmax\_lse}_{\text{full}}) < Threshold$}
                    \State Add Vertical \& Slash Budgets to the config $c$.
                \EndIf
            \EndFor
        \EndFor
    \EndProcedure
\end{algorithmic}
\end{algorithm}

\paragraph{Impact of Sparse Attention on Accuracy}

To demonstrate the necessity of our method of integrating DCA and sparsity refinement, we evaluate Qwen2.5-7B-Instruct-1M on the Needle in a Haystack Test~\citep{niah} with context lengths up to 1 million tokens. We choose this model because smaller models exhibit lower tolerance for information losses due to sparse attention, thereby better highlighting the value of our improvements.

\begin{figure}[!h]
    \centering
    \includegraphics[width=0.85\textwidth]{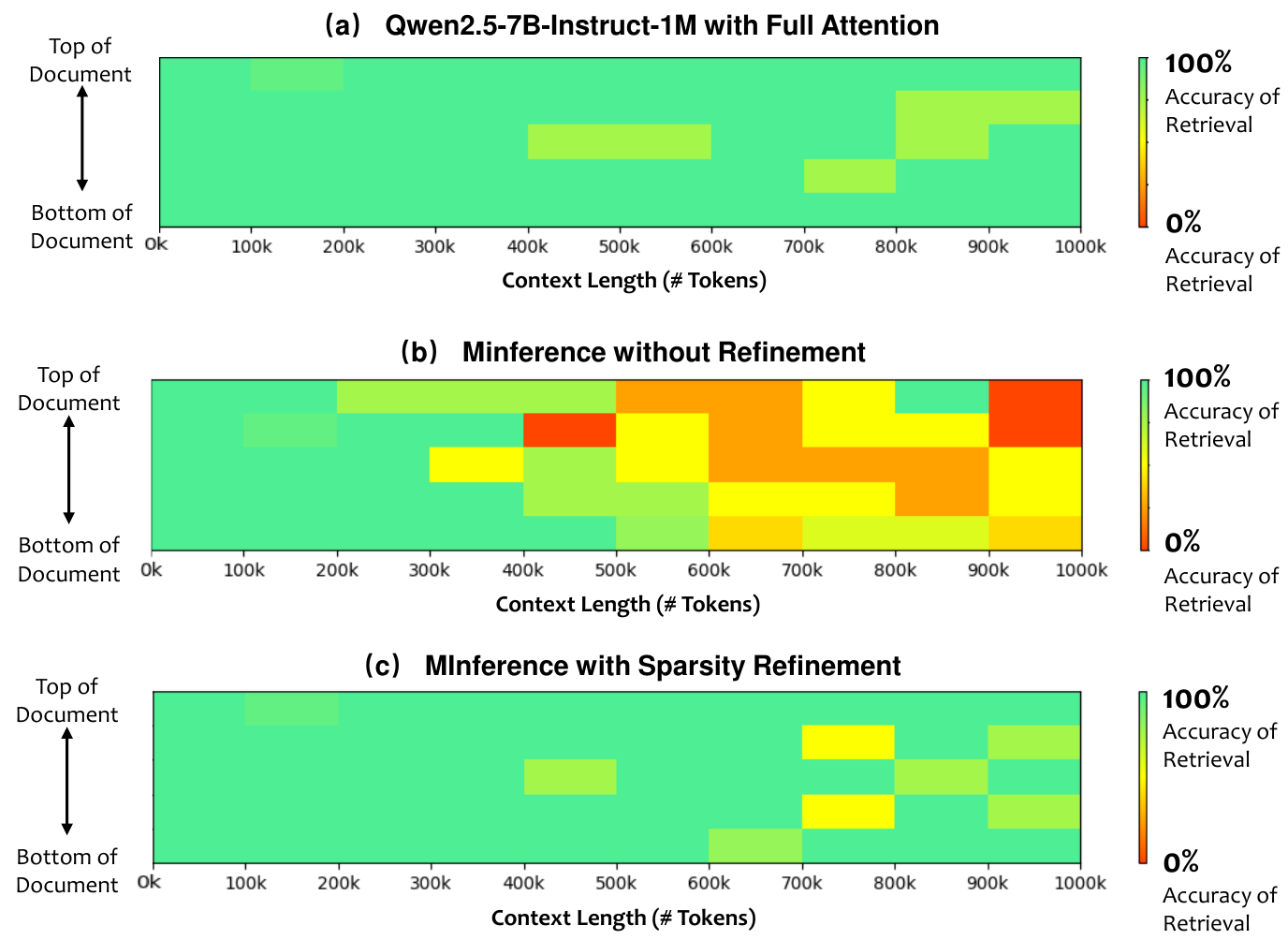}
    \caption{\textbf{Evaluate Qwen2.5-7B-Instruct-1M on Needle in A Haystack with Different Sparsification Configurations.}\label{fig:minference_ablation}}
\end{figure}

As illustrated in Figure \ref{fig:minference_ablation}, Qwen2.5-7B-Instruct-1M with full attention successfully retrieves the majority of needles even in contexts of 1 million tokens. However, using the original MInference method results in a significant performance drop. For context lengths exceeding 400k tokens, the model's retrieval accuracy can fall to 60\% or lower.

After incorporating continuous relative positions to select critical tokens and refining the sparsification configuration, as shown in Figure \ref{fig:minference_ablation}(c), the model recovers most of the performance and maintains about 4 times speedup during the prefilling stage.

\subsection{Inference Engine}

In addition to algorithmic advancements, optimizing the inference engine is essential for enabling LLMs to process long sequence effectively. The API services for Qwen2.5-1M Models are powered by BladeLLM, a high-performance inference engine developed by the Alibaba PAI Engine team. BladeLLM has been specifically optimized for long-sequence prefill and decoding through enhancements in kernel performance, pipeline parallelism, and scheduling algorithms.
To assist the open-source community in efficiently deploying the Qwen models with extended context lengths, several of these optimizations have been open-sourced and will be integrated into vLLM~\citep{vllm}.

\subsubsection{Kernel Optimization}

\paragraph{Sparse Attention Kernel Optimization}

As the context length ($L_c$) increases, the computational complexity of attention ($O(L_c^2)$) and memory access ($O(L_c)$) grow significantly. To tackle this issue, the industry has explored optimization strategies to enhance efficiency through sparsity, such as the Vertical-Slash method by MInference~\citep{jiang2024minference}. However, we observe that the efficiency of the attention kernel after sparsification remains low, still resulting in a significant proportion of the total inference time in end-to-end applications. Therefore, extreme optimization of the sparse attention kernel is even more crucial to fully unleashing the potential of sparsification.

To mitigate the overhead associated with sparse memory access, we implemente multi-stage pipeline parallelism and performed intensive instruction-level optimization when loading sparse KV pairs from global memory. 
Our optimized sparse attention kernel is engineered to leverage the capabilities of various GPU architectures, including NVIDIA's Ampere and Hopper series, AMD's MI300 series, and other hardware platforms.

\begin{figure}[t]
    \centering
    \begin{minipage}[b]{0.53\textwidth}
        \centering
	\includegraphics[width=\linewidth]{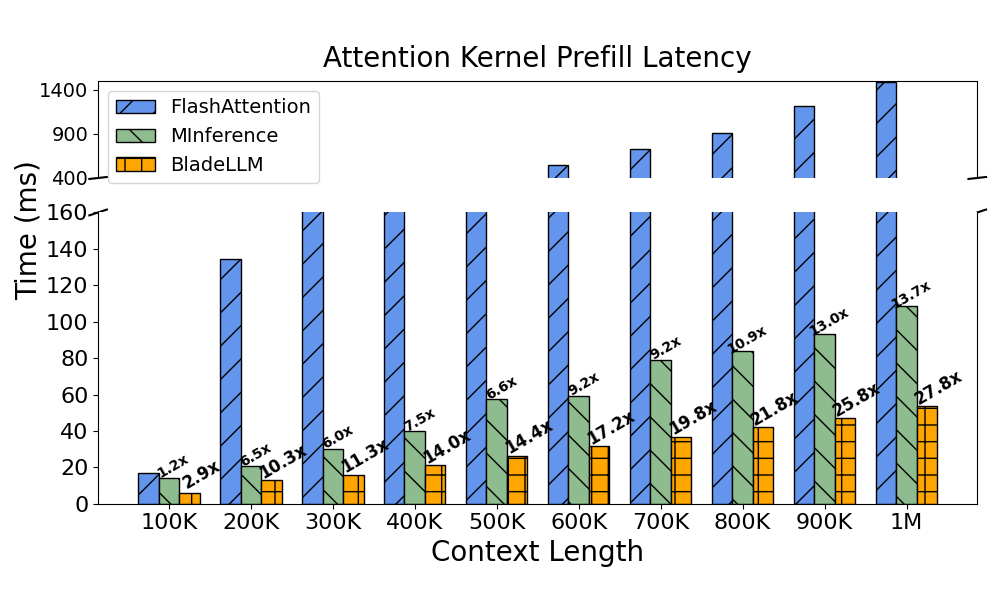}
    \vspace{-1.8em}
	\caption{\textbf{Performance of Sparse Attention Kernels.}}
	\label{fig:bench_spattn}
    \end{minipage}\hfill
    \begin{minipage}[b]{0.47\textwidth}
	\includegraphics[width=\linewidth]{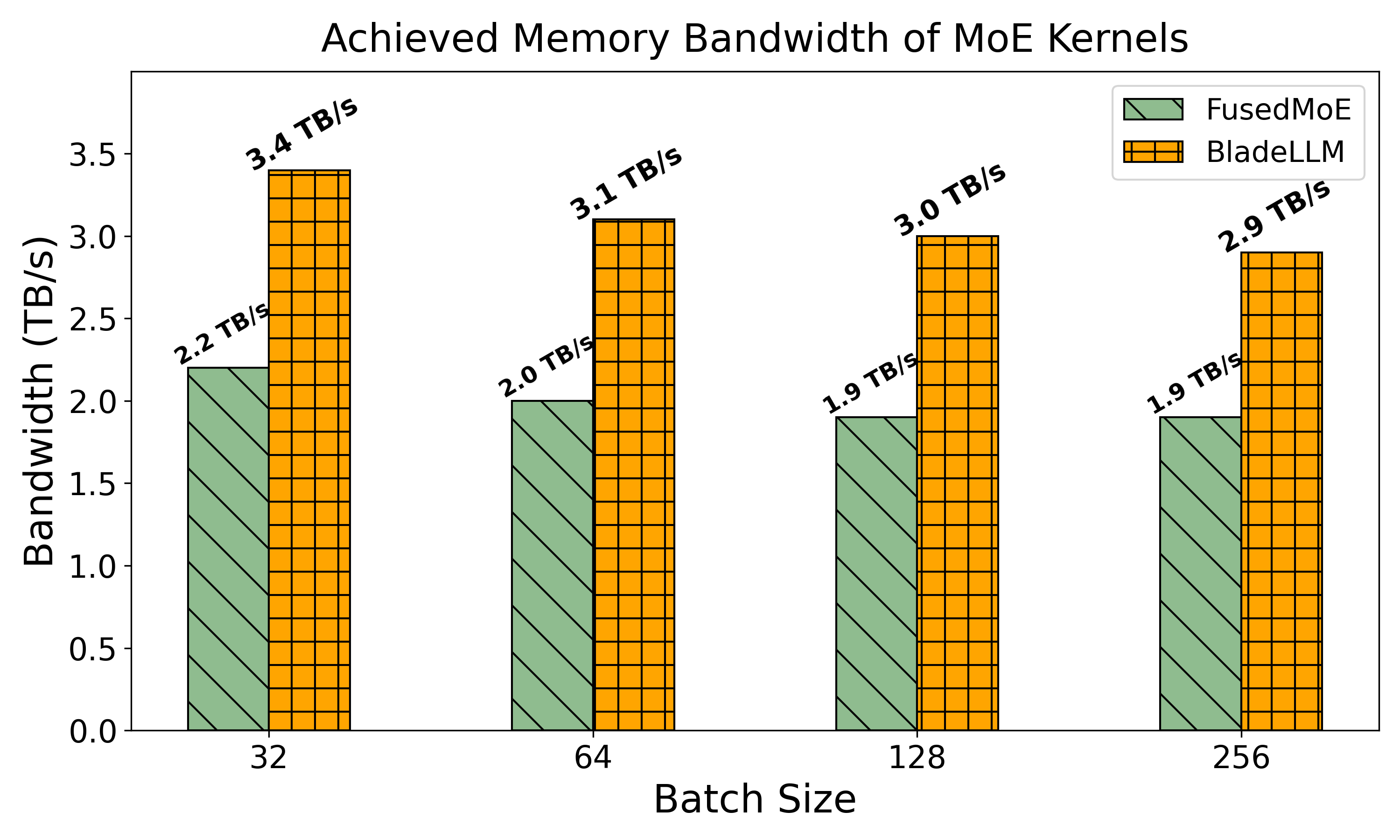}
	\caption{\textbf{Performance of MoE Kernels.}}
	\label{fig:bench_moe}
    \end{minipage}
\end{figure}

Our experiments demonstrate that the optimized kernel in BladeLLM achieves remarkable high computational efficiency across multiple hardware platforms, with a peak FLOPs utilization rate of up to 90\%. As shown in Figure \ref{fig:bench_spattn}, on the A100 GPU, under a 1 million token context, MInference exhibits 13.7x speedup compared to FlashAttention, while BladeLLM achieves 27.8x speedup under the same sparsity configuration.

\paragraph{MoE Kernel Optimization}

Mixture-of-Experts (MoE) models, characterized by their sparse activation of parameters and outstanding accuracy, are particularly well-suited for large-scale deployment. In optimizing the performance of our MoE model, Qwen2.5-Turbo, we have identified that decoding performance is significantly influenced by memory access speed. Specifically, during the decoding phase, when handling batch sizes of 32 or greater, the access to large model parameters in each decoding iteration becomes a critical bottleneck for the overall efficiency of MOE layers. Therefore, enhancing memory access efficiency within the MoE kernel is crucial to achieving peak decoding performance.

BladeLLM enhances the efficiency of MoE kernels through a variety of optimization techniques, including improved Tensor Core utilization specifically tailored for memory-bound scenarios and fine-grained warp specialization. On the H20 GPU, these optimizations achieve peak memory access efficiency of 3.4 TB/s, representing a 55\% improvement over the FusedMoE kernels in vLLM. Performance results across different batch sizes are illustrated in Figure \ref{fig:bench_moe}.

\subsubsection{Dynamic Chunked Pipeline Parallelism}

\begin{figure}[h]
    \centering
     \begin{subfigure}[b]{0.49\textwidth}
         \centering
         \includegraphics[height=0.08\textheight,width=\textwidth]{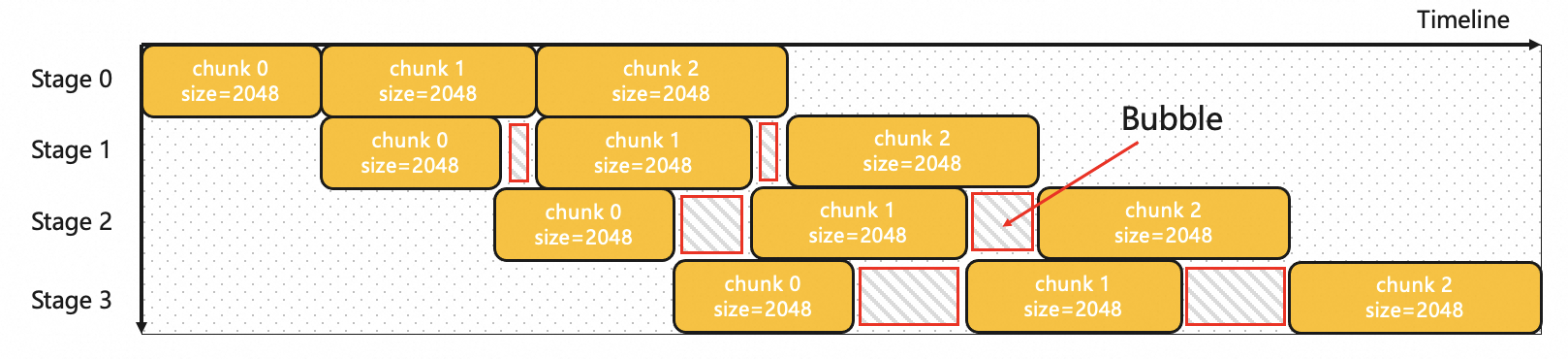}
         \caption{Pipeline Bubbles in Chunked Prefilling}
         \label{subfig:pp_bubble}
     \end{subfigure}
    ~
     \begin{subfigure}[b]{0.49\textwidth}
         \centering
         \includegraphics[height=0.08\textheight,width=\textwidth]{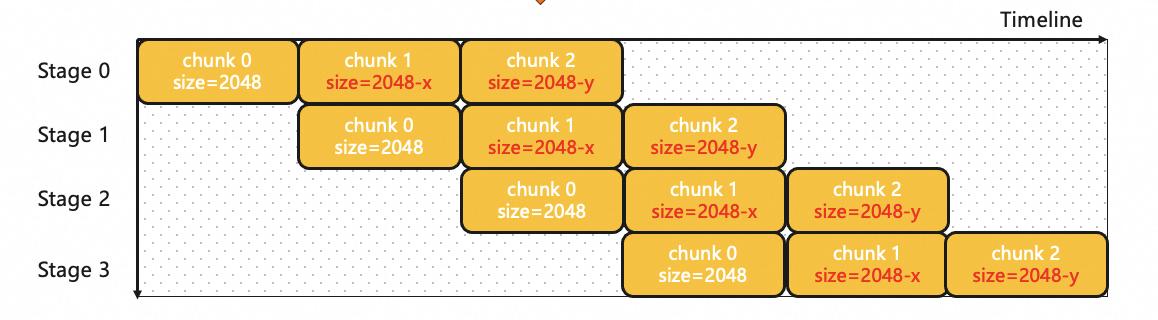}
         \caption{Dynamic Chunked Pipeline Parallelism}
         \label{subfig:dcpp}
     \end{subfigure}
    
    \caption{\textbf{Optimization of Pipeline Parallelism in BladeLLM.}\label{fig:pp}}
\end{figure}

Pipeline parallelism is a technique that divides the model into multiple segments, allowing different parts of the model to be processed concurrently. This approach significantly reduces communication volume compared to tensor parallelism, and its kernel execution efficiency is enhanced due to more intensive operations. In recent industry practices, segmented pipeline parallelism (Chunked Pipeline Parallelism) has been employed to accelerate the prefilling phase.
Nevertheless, in scenarios involving extensive context lengths, we have identified an issue: variations in history length (i.e., the length of the past KV Cache) across different chunks lead to substantial disparities in attention computation time. This discrepancy results in numerous pipeline bubbles, as depicted in Figure \ref{fig:pp}(a).

BladeLLM employs Dynamic Chunked Pipeline Parallelism (DCPP) for long-context prefilling, dynamically adjusting the chunk size based on the computation complexity of the attention kernel to ensure that the execution time of each chunk is as equal as possible, thereby minimizing pipeline bubbles, as shown in Figure \ref{fig:pp}(b).

\subsubsection{Scheduling}

A typical LLM inference engine can be divided into the following components:

\begin{itemize}
    \item \textbf{API Server}: Responsible for receiving requests and sending responses.
    \item \textbf{Scheduler}: Responsible for request scheduling, KV Cache Block allocation, etc.
    \item \textbf{Model Runner}: Responsible for model computation and sampling.
    \item \textbf{Decoder}: Responsible for converting sampled Token IDs into texts for output.
\end{itemize}

\begin{figure}
	\centering
    \begin{subfigure}[b]{0.6\textwidth}
    \centering
        \includegraphics[width=1\textwidth]{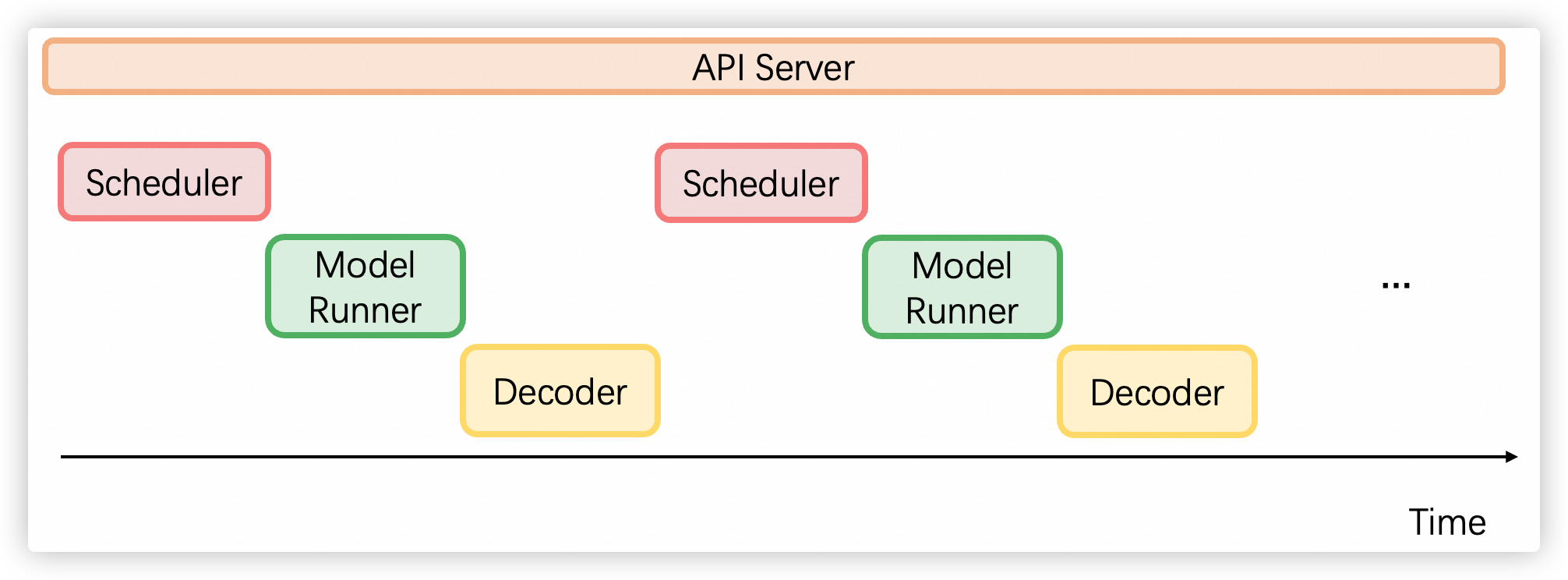}
        \caption{Serial Execution Pipeline of Early Mainstream Inference Engines}
    \end{subfigure}
    \vspace{1em}
    
    \begin{subfigure}[b]{0.6\textwidth}
    \centering
        \includegraphics[width=1\textwidth]{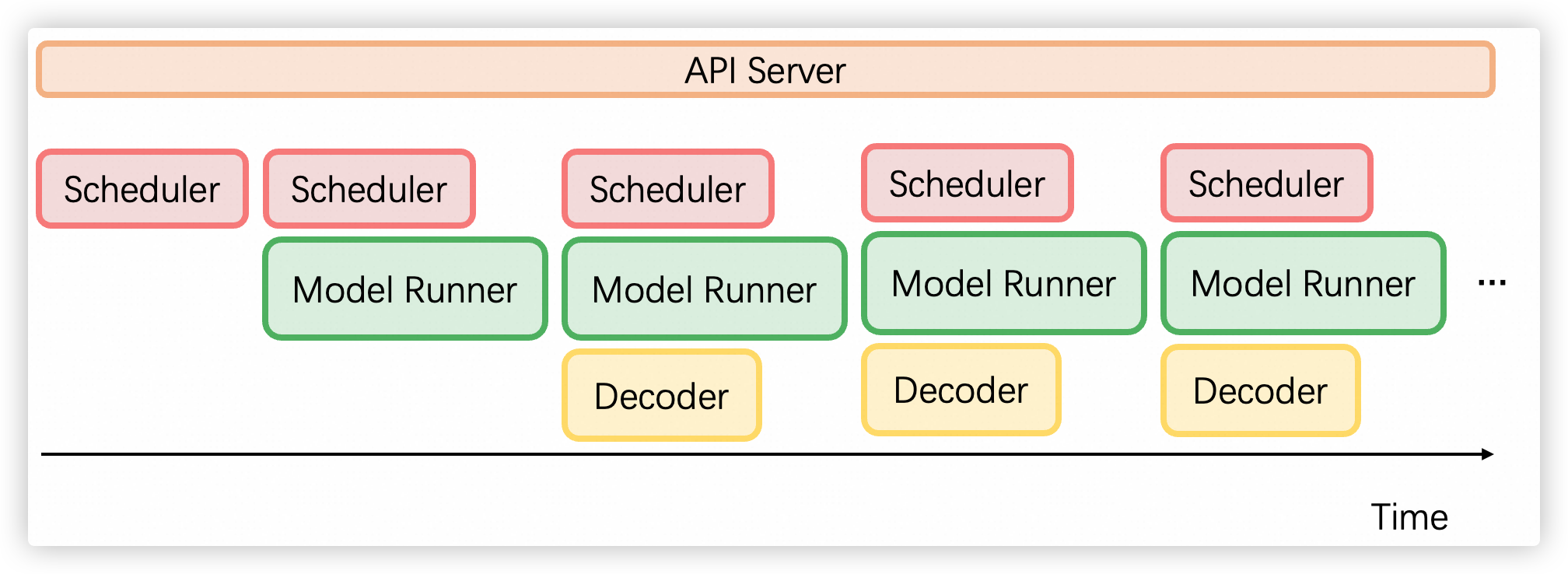}
        \caption{Totally Asynchronous Generator (TAG)}
    \end{subfigure}
	\caption{\textbf{Scheduling Optimization in BladeLLM}.}
	\label{fig:tag}
\end{figure}

In early mainstream inference engines, the Scheduler, Model Runner, and Decoder operated in a serial manner, as shown in Figure \ref{fig:tag}(a). In this setup, non-GPU operations such as the Scheduler and Decoder occupied a significant portion of the decode time, leading to lower GPU utilization.

To address these issues, BladeLLM has implemented a fully asynchronous LLM inference architecture called Totally Asynchronous Generator (TAG), as shown in Figure \ref{fig:tag}(b). Specifically, the three components are handled by three separate processes, where no synchronization is required among them.

\begin{enumerate}
    \item \textbf{Scheduler}: Allocates KV Cache for the next Model Runner step based on anticipated tokens (usually 1, but can be more in speculative sampling) without waiting for previous results of Model Runner.
    \item \textbf{Model Runner}: Retrieves requests from the queue allocated by the Scheduler and processes them. After processing, it places the sampled token IDs directly into the Decoder's queue and continues with the next computation step.
    \item \textbf{Decoder}: Asynchronously retrieves token IDs from the queue, converts them to text, and sends them to the API Server.
\end{enumerate}

Furthermore, BladeLLM employs shared memory across its components to further reduce inter-process communication overhead. Through these methods, BladeLLM significantly reduces overhead in non-GPU stages of the inference engines, substantially enhancing decoding efficiency.

\section{Evaluation}

To comprehensively evaluate the performance of the Qwen2.5-1M series models, we will begin by assessing their capabilities in long-context tasks, highlighting the significant improvements achieved through our specialized long-context optimizations. Next, we will examine their performance in short-context tasks and compare these results with those of the 128k version. Finally, we will demonstrate the inference speed of the models.

\subsection{Long Context Benchmarks}

We first evaluate the Qwen2.5-1M series of models on the Passkey Retrieval task with a context length of 1 million tokens. As illustrated in Figure \ref{fig:cover}, both the Qwen2.5-14B-Instruct-1M and Qwen2.5-Turbo models achieved perfect accuracy, successfully identifying all hidden numbers within the contexts up to 1 million tokens. The smaller Qwen2.5-7B-Instruct-1M model also performed admirably, with only a few minor errors. These results highlight the robust retrieval capabilities of the Qwen2.5-1M models when processing extensive 1 million token contexts.

For more advanced tasks, we utilize three benchmarks to evaluate long-context capabilities:

\begin{itemize}
    \item RULER~\citep{hsieh2024ruler}: An extension of the needle-in-a-haystack task, RULER challenges models to find multiple "needles" or answer multiple questions within irrelevant contexts, or to identify the most or least frequent words in the text. The maximum data length is 128K tokens.
    \item LV-Eval~\citep{yuan2024lveval}: This benchmark tests a model's ability to understand numerous evidence fragments simultaneously. We have refined the evaluation metrics from the original LV-Eval to avoid false negatives caused by overly strict matching rules. The maximum data length is 256K tokens.
    \item Longbench-Chat~\citep{longalign2024bai}: A dataset for evaluating human preference alignment in long-context tasks. The maximum data length is 100K tokens.
\end{itemize}

For baselines, we choose GLM-9B-Chat-1M~\citep{chatglm4}, Llama-3-8B-Instruct-Gradient-1048k~\citep{gradientlongcontextllama3}, Llama-3.1-70B-Instruct, GPT-4o-mini, and GPT-4o.

The results from the three benchmarks are detailed in Tables \ref{tab:ruler} and \ref{tab:lveval}. It is evident that the Qwen2.5-1M series models significantly outperform their 128k counterparts in most long-context tasks, particularly for sequences exceeding 64k in length. On the RULER dataset, all models in the Qwen2.5-1M series even surpasses GPT-4, highlighting their exceptional capability in long-context retrieval tasks.

Notably, the Qwen2.5-14B-Instruct-1M model achieved an accuracy of 92.2 on 128k sequences, marking the first time any model in Qwen2.5 series surpassed the 90-point threshold. Additionally, it consistently outperforms GPT-4o-mini across multiple datasets, offering a robust open-source alternative for long-context tasks.

Qwen2.5-Turbo's performance is positioned between that of the Qwen2.5-7B-Instruct-1M and Qwen2.5-14B-Instruct-1M models in the long-context benchmarks. It delivers faster inference speeds and lower costs, making it a more cost-effective and efficient option.

Qwen2.5-72B-Instruct, despite being trained on sequences limited to 32k tokens, consistently outperformed the Qwen2.5-14B-Instruct-1M model across all sequence lengths in the LV-Eval benchmark when augmented with our length extrapolation method, DCA+YaRN. This result underscores the substantial value of the length extrapolation technique while also highlighting the inherent advantages of larger models in managing complex long-context tasks.

\begin{table}[tbp]
\centering
\caption{\textbf{Performance of Qwen2.5 Models on RULER.} \textit{DCA+YaRN} does not change the model behavior within its training length.}
\label{tab:ruler}
\small
\begin{tabular}{@{}llrlllllll@{}}
\toprule
\multirow{2}{*}{\bf Model} & & \bf \multirow{2}[2]{*}{\makecell{\bf Claimed \\ Length}} & \multicolumn{6}{c}{\bf RULER}  \\ 
\cmidrule{4-10}
 & & & \bf Avg.  & \bf 4K   & \bf 8K    & \bf 16K  & \bf 32K  & \bf 64K   & \bf 128K  \\ \midrule
 \multicolumn{2}{@{}l}{GLM4-9b-Chat-1M} & 1M         & 89.9   & 94.7   & 92.8            & 92.1   & 89.9   & 86.7          & 83.1             \\
\multicolumn{2}{@{}l}{Llama-3-8B-Instruct-Gradient-1048k} & 1M         & 88.3   & 95.5   & 93.8            & 91.6   & 87.4   & 84.7          & 77.0      \\
\multicolumn{2}{@{}l}{Llama-3.1-70B-Instruct} & 128K & 89.6 & 96.5 & 95.8 &  95.4 & 94.8  & 88.4 &  66.6 \\
\multicolumn{2}{@{}l}{GPT-4o-mini}    & 128K       & 87.3   & 95.0   & 92.9            & 92.7   & 90.2   & 87.6          & 65.8                \\ 
\multicolumn{2}{@{}l}{GPT-4}        & 128K     & 91.6 & 96.6 & 96.3 & 95.2 & 93.2 & 87.0          & 81.2              \\ 
\midrule
\multirow{2}{*}{\textbf{Qwen2.5-32B-Instruct}} & RoPE & 32K & {88.0} & \multirow{2}{*}{96.9} & \multirow{2}{*}{97.1} & \multirow{2}{*}{95.5} & \multirow{2}{*}{95.5} & {85.3} & {57.7}\\
 & DCA+YaRN & 128K & \bf 92.9 &  &  &  &  & \bf 90.3 & \bf 82.0  \\
\multirow{2}{*}{\textbf{Qwen2.5-72B-Instruct}} & RoPE & 32K &  {90.8} & \multirow{2}{*}{\underline{97.7}} & \multirow{2}{*}{\underline{97.2}} & \multirow{2}{*}{\underline{97.7}} & \multirow{2}{*}{\underline{96.5}} &  {88.5} & {67.0} \\
& DCA+YaRN & 128K & \bf95.1 &  &  &  &  & \bf 93.0 & \bf 88.4 \\
\midrule
\multirow{2}{*}{\textbf{Qwen2.5-7B-Instruct}} & RoPE & 32K & {80.1} & \multirow{2}{*}{96.7} & \multirow{2}{*}{95.1} & \multirow{2}{*}{\bf93.7} & \multirow{2}{*}{89.4} &{74.5}&{31.4} \\
 & DCA+YaRN & 128K & 85.4 &  &  &  &  &82.3&55.1\\
\textbf{Qwen2.5-7B-Instruct-1M} & RoPE / DCA+YaRN & 1M & \bf 91.8 & \bf 96.8 & \bf 95.3 &93.0 & \bf 91.1& \bf 90.4 & \bf 84.4\\
\midrule
\multirow{2}{*}{\textbf{Qwen2.5-14B-Instruct}} & RoPE & 32K & {86.5} & \multirow{2}{*}{\underline{\bf97.7}} & \multirow{2}{*}{96.8} & \multirow{2}{*}{\bf95.9} & \multirow{2}{*}{93.4} & {82.3} & {53.0} \\
 & DCA+YaRN & 128K & 91.4 &  & &  & & 86.7 & 78.1 \\
\textbf{Qwen2.5-14B-Instruct-1M} & RoPE / DCA+YaRN & 1M & \bf \underline{95.7} & 97.5 & \bf 97.1 & 94.6 & \bf 94.9& \bf \underline{94.9} & \bf \underline{92.2}\\
\midrule
\textbf{Qwen2.5-Turbo} & RoPE / DCA+YaRN & 1M & {93.1} & {97.5} & 95.7 & {95.5} & {94.8} & {90.8} & {84.5} \\                 
\bottomrule
\end{tabular}
\end{table}

\begin{table}[tbp]
\caption{\textbf{Performance of Qwen2.5 Models on LV-Eval and LongBench-Chat.} \textit{DCA+YaRN} does not change the model behavior within its training length.}
\label{tab:lveval}
\small
\centering
\begin{tabular}{@{}llrrrrrrc@{}}
\toprule
\multirow{2}{*}{\bf Model}  & & \bf \multirow{2}[2]{*}{\makecell{\bf Claimed \\ Length}} & \multicolumn{5}{c}{\bf LV-Eval} & \multirow{2}[2]{*}{\makecell{\bf LongBench-\\ \bf Chat }} \\ 
\cmidrule{4-8}
 & \bf  &  & \bf 16K   & \bf 32K    & \bf 64K  & \bf 128K  & \bf 256K    \\ \midrule
\multicolumn{2}{@{}l}{GLM4-9B-Chat-1M}                    & 1M     &    46.4	&43.2	&42.9	&40.4	&37.0	&7.82                      \\
\multicolumn{2}{@{}l}{Llama-3-8B-Instruct-Gradient-1048k} & 1M          &   31.7	 &31.8	 & 28.8	 & 26.3	& 21.1 &	6.20            \\
\multicolumn{2}{@{}l}{Llama-3.1-70B-Instruct} & 128K & 48.6 & 47.4 & 42.9 & 26.2 & N/A & 6.80 \\
\multicolumn{2}{@{}l}{GPT-4o-mini}                       & 128K    & 52.9	& 48.1	& 46.0	& 40.7	& N/A	& 8.48            \\ 
\midrule
\multirow{2}{*}{\textbf{Qwen2.5-32B-Instruct}} & RoPE & 32K & \multirow{2}{*}{56.0} & \multirow{2}{*}{53.6} & {40.1} & {20.5} & {0.7} & {-}\\
& DCA+YaRN & 128K &  &  & \bf 48.8 & \bf 45.3 & \bf 41.0 & 8.70 \\
\multirow{2}{*}{\textbf{Qwen2.5-72B-Instruct}} & RoPE & 32K & \multirow{2}{*}{\underline{60.4}} & \multirow{2}{*}{\underline{57.5}} & {47.4} & {27.0} & {2.4} & {-} \\
& DCA+YaRN & 128K &  &  &  \bf\underline{53.9} & \bf\underline{50.9} & \bf\underline{45.2} & {8.72} \\
\midrule
\multirow{2}{*}{\textbf{Qwen2.5-7B-Instruct}} & RoPE  & 32K & \multirow{2}{*}{\bf 55.9}&\multirow{2}{*}{\bf 49.7}&{33.1}&{13.6}&{0.5} & {-}\\ 
& DCA+YaRN  & 128K & & &48.0 & 41.1 & 36.9 & 7.42\\
\textbf{Qwen2.5-7B-Instruct-1M} & RoPE / DCA+YaRN & 1M & 52.5 & 49.4 & \bf 48.6 & \bf 48.3 & \bf 42.7 & \bf 8.08 \\
\midrule
\multirow{2}{*}{\textbf{Qwen2.5-14B-Instruct}} & RoPE & 32K & \multirow{2}{*}{53.0} & \multirow{2}{*}{50.8} & {37.0} & {18.4} & {0.8} & {-}\\
& DCA+YaRN & 128K &  &  & 46.8 & 43.6 & 39.4 & 8.04  \\
\textbf{Qwen2.5-14B-Instruct-1M} & RoPE / DCA+YaRN & 1M & \bf 54.5 & \bf 53.5 & \bf 50.1 & \bf 47.6 & \bf 43.3 & \underline{\bf 8.76} \\
\midrule
\textbf{Qwen2.5-Turbo}        & RoPE / DCA+YaRN    & 1M     & 53.4	& 50.0	&45.4 	& 43.9	& 38.0&	8.34        \\   
\bottomrule
\end{tabular}
\end{table}

\subsection{Short Context Benchmarks}

In addition to performance improvements in long-context tasks, we conducted a comprehensive comparison between Qwen2.5-1M and its 128k counterpart on short-context tasks.

We selected widely used benchmarks targeting natural language understanding, coding, mathematics, and reasoning.
For general evaluation, we utilized MMLU-Pro~\citep{mmlupro}, MMLU-redux~\citep{mmluredux}, and LiveBench 0831~\citep{livebench}. For science and mathematics, we evaluated the models on GPQA~\citep{gpqa}, GSM8K~\citep{gsm8k}, and MATH~\citep{math}. In coding, we assessed performance using HumanEval~\citep{humaneval}, MBPP~\citep{mbpp}, MultiPL-E~\citep{multiple}, and LiveCodeBench 2305-2409~\citep{livecodebench}. Additionally, we measured instruction-following capabilities using IFEval~\citep{ifeval}, where we report results for the strict prompt-level accuracy.
To further evaluate human preference alignment and instruction-following performance, we assessed the models on MT-Bench \citep{mtbench} and Arena-Hard \citep{arena-hard}.

\begin{table}[t]
\centering
\caption{\textbf{Performance of Qwen2.5-7/14B-Instruct with the 1M versions and Qwen2.5-Turbo.}}

\label{tab:short_results}
\small
\setlength{\tabcolsep}{1.4pt}

\begin{tabular}{lcccccc}
\toprule

\textbf{Datasets} & \textbf{GPT4o-mini} & \textbf{Qwen2.5-7B}  & \textbf{Qwen2.5-7B-1M} & \textbf{Qwen2.5-14B} & \textbf{Qwen2.5-14B-1M} & \textbf{Qwen2.5-Turbo} \\

\midrule

\multicolumn{7}{c}{\textit{General Tasks}} \\

\midrule

MMLU-Pro & 63.1 & 56.3 & 54.3 & 63.7 & 63.3 & \bf64.5\\

MMLU-redux & 81.5 & 75.4 & 74.8 & 80.0 & 80.7 & \bf81.7\\

LiveBench {\tiny0831} & 31.1 & 35.9 & 35.2 & 44.4 & \bf44.6 & 42.3\\

\midrule

\multicolumn{7}{c}{\textit{Mathematics \& Science Tasks}} \\

\midrule

GPQA & 40.2 & 36.4 & 41.4 & \bf45.5 & 39.9 & 42.3\\

MATH & 70.2 & 75.5 & 72.9 & 80.0 & 79.5 & \bf81.1\\

GSM8K & 93.2 & 91.6 & 91.7 & \bf94.8 & \bf94.8 & 93.8\\

\midrule

\multicolumn{7}{c}{\textit{Coding Tasks}} \\

\midrule

HumanEval & 88.4 & 84.8 & 86.0 & 83.5 & \bf88.4 & 86.6\\

MBPP & \bf85.7 & 79.2 & 75.9 & 82.0 & 80.2 & 82.8\\

MultiPL-E & 75.0 & 70.4 & 72.4 & 72.8 & \bf77.1 & 73.7\\

LiveCodeBench & 40.7 & 28.7 & 28.0 & \bf42.6 & 38.6 & 37.8\\

\midrule

\multicolumn{7}{c}{\textit{Alignment Tasks}} \\

\midrule

IFEval & 80.4 & 71.2 & 73.0 & 81.0 & \bf84.3 & 76.3\\

Arena-Hard & \bf74.9 & 52.0 & 48.1 & 68.3 & 70.2 & 67.1\\

MTbench & - & 8.75 & 8.30  & 8.88 & \bf8.89 & 8.81\\

\bottomrule

\end{tabular}

\end{table}

As shown in Table \ref{tab:short_results}, Qwen2.5-7B-Instruct-1M and Qwen2.5-14B-Instruct-1M maintain performance on short text tasks that is similar to the 128k versions, ensuring that their fundamental capabilities have not been compromised by the addition of long-sequence processing abilities. Compared to GPT-4o-mini, both Qwen2.5-14B-Instruct-1M and Qwen2.5-Turbo achieve similar performance on short text tasks while supporting a context length that is eight times longer.

\subsection{Speed Comparison}

\begin{figure}[t]
    \centering
    \includegraphics[width=1\textwidth]{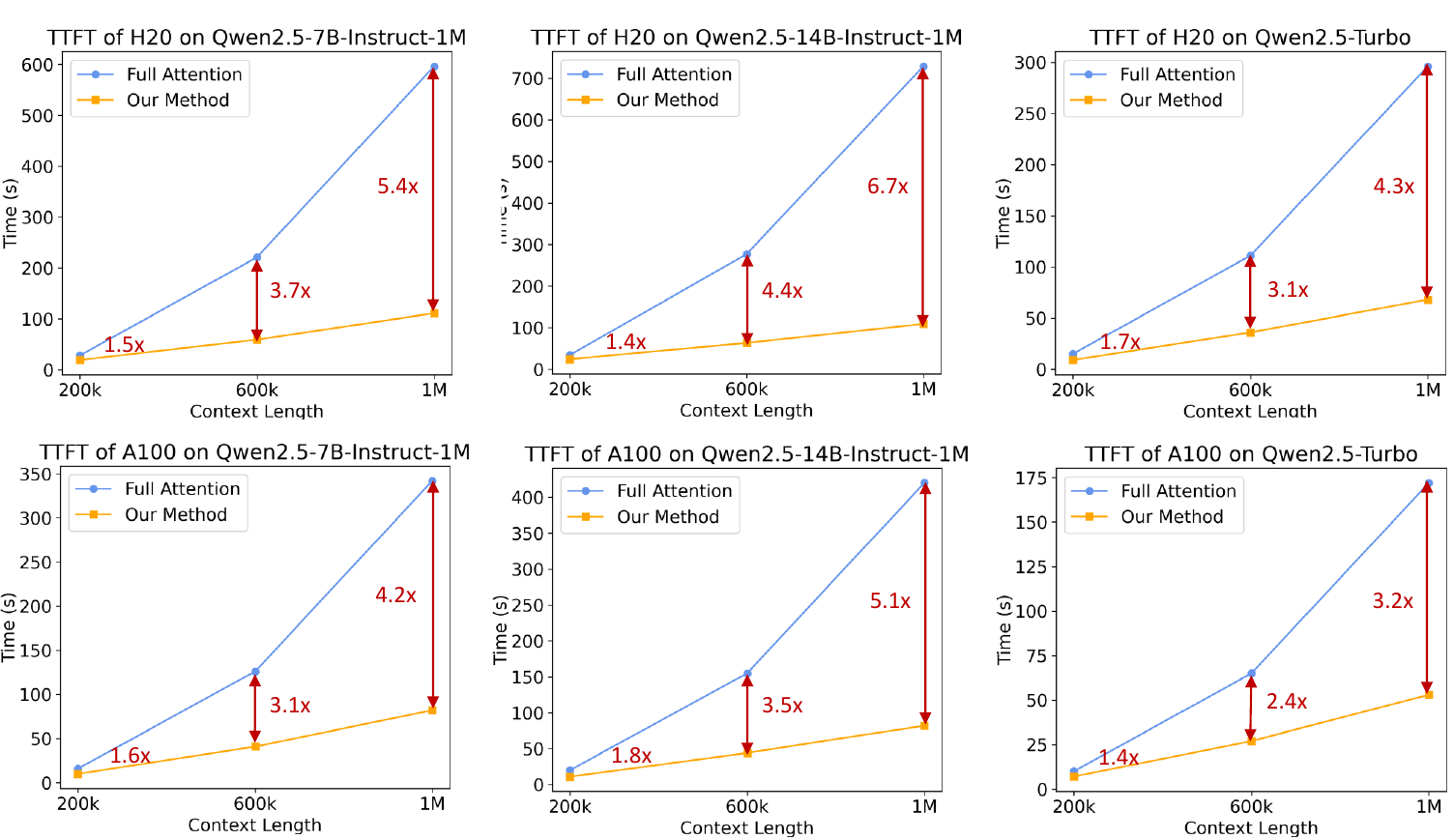}
    \caption{\textbf{TTFT (Time to First Token) of Qwen2.5-7B-Instruct-1M, Qwen2.5-14B-Instruct-1M, Qwen2.5-Turbo on H20 and A100 GPUs}.}
     \label{fig:speed}
\end{figure}

To demonstrate the acceleration of our final solution on processing long sequences, we evaluate the Time to First Token (TTFT) for different context lengths on Nvidia H20 and A100 GPUs. The experimental configurations are as follows:

\begin{itemize}
    \item For Qwen2.5-14B-Instruct-1M and Qwen2.5-Turbo, we employed tensor parallelism with 8-way partitioning.
    \item For Qwen2.5-7B-Instruct-1M, due to the constraints imposed by the Grouped Query Attention mechanism, we utilized tensor parallelism with 4-way partitioning.
\end{itemize}

In all experiments, we use a batch size of 1.

As illustrated in Figure \ref{fig:speed}, our method, enhanced with sparse attention and optimized inference engines, achieves a 3.2 to 6.7 times speedup when processing a context length of 1M across various model sizes and devices.
For instance, on the H20 GPUs, the Qwen2.5-14B-Instruct-1M model reduces inference time from 12.2 minutes (with full attention) to just 109 seconds. Similarly, the Qwen2.5-Turbo model decreases its processing time from 4.9 minutes to only 68 seconds. These improvements significantly reduce user waiting times for long-sequence tasks.

Compared to the open-source Qwen2.5-1M models, Qwen2.5-Turbo excels in short tasks and achieves competitive results on long-context tasks, while delivering shorter processing times and lower costs.  Consequently, it offers an excellent balance of performance and efficiency, making it highly cost-effective.

\section{Conclusion}
\label{sec:conclusion}

In this technical report, we introduce the Qwen2.5-1M series of models, which includes the open-source models Qwen2.5-7B-Instruct-1M and Qwen2.5-14B-Instruct-1M, as well as the API-accessible model Qwen2.5-Turbo. We detail how these models were developed by extending the Qwen2.5 Base model through long-context pre-training and post-training. We introduce techniques such as data synthesis and progressive training to improve training effectiveness with lower costs.

Beyond training, efficient inference and deployment present significant challenges for long-context models. We have implemented several optimizations, including training-free length extrapolation methods, sparse attention mechanisms, and inference engine enhancements. These optimizations substantially improve the efficiency and reduce the operational costs of running long-sequence models, making practical applications more feasible. We have open-sourced several of these optimizations, and firmly believe that this is the most effective way to drive progress in the field.

We recognize that long-context models still have significant potential for improvement. Our focus is on developing models that excel in both short and long-context tasks, ensuring they deliver substantial value in real-world long-context scenarios. We will continue to explore more efficient training strategies, model architectures, and inference methods, making them deployable effectively and perform exceptionally well even in resource-constrained environments. We are confident that these efforts will expand the applicability of long-context models to a much broader range of use cases.

\bibliography{biblio}
\bibliographystyle{colm2024_conference}

\end{document}